\documentclass[letterpaper]{article}
\usepackage{aaai}
\usepackage{times}
\usepackage{helvet}
\usepackage{courier}
\usepackage[pdftex]{graphicx} 
\usepackage{caption}
\usepackage{subcaption} 
\usepackage{enumitem}

\begin{document}   
\title{Super Mario as a String: \\ Platformer Level Generation Via LSTMs}
\author{Adam Summerville and  Michael Mateas \\
Expressive Intelligence Studio \\
Center for Games and Playable Media \\
University of California, Santa Cruz \\
\texttt{asummerv@ucsc.edu , michaelm@soe.ucsc.edu}
}
\maketitle
\begin{figure*}
\includegraphics[width=1.0\textwidth]{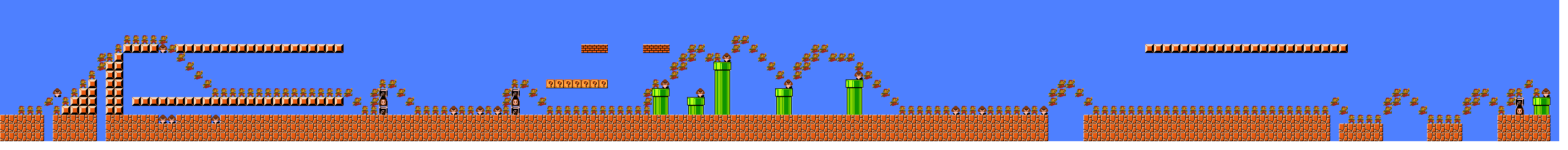}
\end{figure*}
\begin{abstract}
\begin{quote}
The procedural generation of video game levels has existed for at least 30 years, but only recently have machine learning approaches been used to generate levels without specifying the rules for generation.  A number of these have looked at platformer levels as a sequence of characters and performed generation using Markov chains.  In this paper we examine the use of Long Short-Term Memory recurrent neural networks (LSTMs) for the purpose of generating levels trained from a corpus of \textit{Super Mario Brothers } levels.  We analyze a number of different data representations and how the generated levels fit into the space of human authored \textit{Super Mario Brothers} levels. 
\end{quote}
\end{abstract}
\section{INTRODUCTION}

Procedural Content Generation (PCG) of video game levels has existed for many decades, with the earliest known usage coming from \textit{Beneath Apple Manor} \nocite{BENEATH_APPLE_MANOR} in 1978.  As an area of academic study, there have been numerous approaches taken ranging from human-authored rules of tile placement (\cite{MARIOAI}), to constraint solving and planning (\cite{TANAGRA}). These approaches require the designer of the algorithm to specify either the rules of the generation or a method for evaluating the generated levels.  More recently, machine learning approaches have been used so that a designer can train a generator from examples. These have ranged from matrix factorization (\cite{SHAKERALONE}) (\cite{ZELDA}), to a number of different Markov chain approaches (\cite{NGRAMLEVELS})(\cite{SNODGRASS})(\cite{MCMCTS}), to different graph learning approaches (\cite{GUZDIAL})(\cite{MARIOGRAPHGRAMMAR}), to Bayesian approaches (\cite{BAYESZELDA}).  A number of these approaches treat the level as a sequence of characters, and use Markov chains to learn character-to-characters transition probabilities.  These approaches can have good local coherence but struggle with global coherence.  This can be remedied by considering longer and longer histories, but this leads to a rapidly exploding combinatoric space leading to a very sparse set of learned transitions.  

Long Short-Term Memory recurrent neural networks (LSTMs) represent the state of the art in sequence learning.  They have been used for translation tasks (\cite{SEQ_TO_SEQ}), speech understanding (\cite{SPEECH_RECOGNITION}), video captioning (\cite{TRANSLATING_VIDEOS}), and more.  In this paper we present a method for utilizing LSTMs for the purpose of learning and generating levels.  We examine eight different data representations for the levels and compare the generated levels using a number of different evaluation statistics.  

\section{RELATED WORK}

Machine learned platformer level generation has mostly been performed via learning Markov chain transition probabilities.  The transitions in the Markov chains has seen two major approaches, either tile-to-tile transitions or vertical slices of the level.  The vertical slice approach was first used by Dahlskog et al. (\cite{NGRAMLEVELS}) and was later used by Summerville et al. (\cite{MCMCTS}).  In this approach the states of the Markov chain are taken as a vertical slice of tiles and the transitions are learned from slice-to-slice.  This has the benefit of directly encoding vertical structure which is important due to the fact that certain tiles are much more likely to be at the top (like empty tiles) and others are much more likely to be at the bottom (such as the ground).  However, this comes with the severe drawback of removing the ability to generate novel vertical slices. It also has the problem that the state space is much larger than just the number of tiles, making the learned transition space much sparser than just learning tile-to-tile transitions.

Snodgrass and Onta\~{n}\'{o}n (\cite{SNODGRASS,SNODGRASS2}) have used tile-to-tile transitions for the placement of tiles.  Because of this, they do not have the problems of the vertical slice approach, but this brings a host of other issues.  Namely, it can be far too likely for the system to generate sequences of tiles that violate implicit semantic constraints, such as generating ground tiles at the top of the level.  Multiple attempts were made to reduce such problems via learning different chains for varying heights, or learning a hierarchical Markov chain (\cite{HDMC}).
We build on the tile-to-tile approach for the work we report here as it is better able to generalize over a larger collection of levels. With the column based approach it is very easy to imagine that a new level might have a column that has never been seen before and so the system will be unable to generalize, but it is much harder to imagine a single-tile-to-tile transition that has never been seen (unless it is one that should never be seen like an enemy embedded inside a pipe).

The above approaches both have the drawback of no guarantees about playability.  In Snodgrass's work, over half of the levels produced were unable to be completed by a simulated agent (\cite{SNODGRASS}).
Summerville, et al. (\cite{MCMCTS}) used Monte Carlo Tree Search as a way of guiding Markov chain construction of levels to help guarantee a playable level.  This does allow for some authorial input on the part of a designer by letting them change the reward function to create levels with different features (e.g. more/less gaps), at the cost of biasing the generation away from the learned transition probabilities in ways that might not be easily understood or resolved.  
Despite the different applications of Markov chains, they all suffer from the same main drawback, that there is no good way to have a long memory in the chain without needing massive amounts of data to fill the transition table.  This leads to good local coherence but poor global coherence, in that they can reliably pick the next tile in a way that makes sense, but have difficulty handling larger structures or patterns.

The work of Guzdial and Riedl (\cite{GUZDIAL}) used clustering to find latent groupings and then learned relative placements of tile blocks in Mario levels.  This has the benefit of learning structure over much longer ranges than is possible in the Markov chain examples. While the work of Dahlskog was able to use 3 tiles worth of history Guzdial's work considered tile distances up to 18 tiles in distance.  A similar graph approach was used by Londo\~{n}o and Missura (\cite{MARIOGRAPHGRAMMAR}), but learned a graph grammar instead.  Their work is unique in platformer level generation in that they used semantic connections such as tile X is reachable from tile Y instead of just physical connections.

The only other known Neural Network approach is that of Hoover et al. \cite{HOOVER_SMB}. They used a single level at a time as training input and tried to predict the height of a specific block type given the height of that block type in the previous 2 time steps.  Extensions tried to predict the height of a given block type given the other block types, the idea being that a user could draw in portions of the level and allow the system to fill in the rest.

Procedural generation via neural networks has most been focused on image and text generation.  Both have typically been byproducts of trying to train a network for some sort of classification process.  Google's Inceptionism (\cite{INCEPTIONISM}) took the product of training an image classifier and manipulated images to better match the set classification class, such as dogs.  Facebook's Eyescream had image generation as the goal (\cite{EYESCREAM}) and used adversarial networks, one network trying to generate images and the other trying to determine if what it was seeing was generated or real.

Text generation from neural networks has been done by numerous people for various goals such as translation (\cite{SEQ_TO_SEQ}), text classification (\cite{GenSeq}), and even trying to determine the output of code simply by reading the source code (\cite{LEARNING_TO_EXECUTE}) and then writing what it believes to be the output.

The work of Graves (\cite{GenSeq}) went beyond text generation and used sequences of stylus locations and pressure to learn calligraphic character writing. Similarly, the DRAW system (\cite{DRAW}) used LSTMs to generate address signs learned from Google street view  by using an attentional system to determine where to focus and then using that as the sequencing of where to place a virtual paint brush and what color to paint, e.g. it learns a sequence of $(x,y,r,g,b)$.  

\section{LSTM SEQUENCE GENERATION}

In this section we will first cover how LSTM Recurrent Neural Networks operate.  We will then cover the basics of our data specification as well as the variants we used. 

\subsection{Recurrent Neural Network Architecture}

\begin{figure*}[ht]
\centering
    \includegraphics[width=0.4\textwidth]{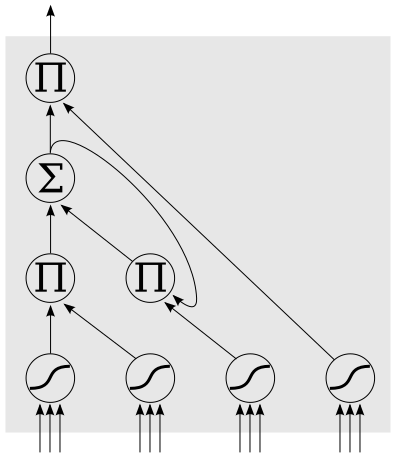}
    \caption{Graphical depiction of an LSTM block. }
    \label{fig:LSTM}
\end{figure*}

Recurrent Neural Networks (RNNs) have been used for a number of different sequence based tasks.  They operate in a manner similar to standard neural networks, i.e. they are trained on data and errors are back-propagated to learn weight vectors. However, in an RNN the edge vectors are not just connected from input to hidden layers to output, they are also connected from a node to itself across time.  This means that back-propagation occurs not just between different nodes, but also across time steps. However, a common problem with standard RNNs is known as ``the vanishing gradient problem'' (\cite{VANISHING_GRADIENTS}), where the gradient of the errors very quickly dissipates across time.  While different in why it occurs, the end result is similar to the problem found with Markov chains, i.e. good local coherence but poor global coherence.   

LSTMs are a neural network topology first put forth by Hochreiter  and Shmidhuber (\cite{LSTM}) for the purposes of eliminating the vanishing gradient problem.  LSTMs work to solve that problem by introducing additional nodes that act as a memory mechanism, telling the network when to remember and when to forget.  A standard LSTM architecture can be seen in figure 1.  At the bottom
are nodes that operate as in a standard neural network, i.e. inputs come in, are multiplied by weights, summed, and that is passed through some sort of non-linear function (most commonly a sigmoid function such as the hyperbolic tangent) as signified by the S-shaped function.  The nodes with $\sum$ simply sum their inputs with no non-linear activation, and the nodes with $\prod$ simply take the product of their inputs.  With that in mind, the left-most node on the bottom can be thought of as the \textit{input} to an LSTM block (although for all purposes it is interchangeable with the node second from the left).  The node second from the left is typically thought of as the \textit{input gate}, since it is multiplied with the input, allowing the input through when it is close to 1, and not allowing the input in when it is close to 0.  Similarly, the right-most node on the bottom acts as a corresponding \textit{output gate}, determining when the value of the LSTM block should be output to higher layers.  The node with the $\sum$ acts as the \textit{memory}, summing linearly and as such not decaying through time.  It feeds back on itself by being multiplied with the second from the right node, which acts as the \textit{forget gate}, telling the memory layer when it should drop whatever it was storing.

\begin{figure*}[ht]
\centering
    \includegraphics[width=0.7\textwidth]{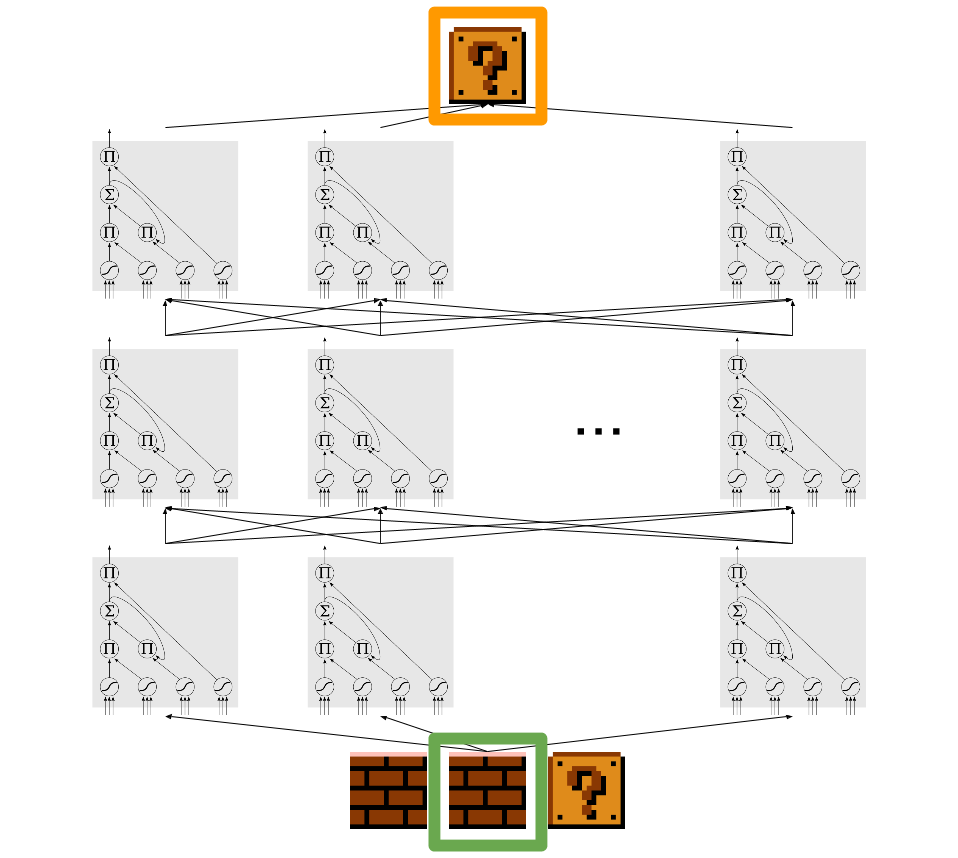}
    \caption{ Graphical depiction of our chosen architecture.  The green box at the bottom represents the current tile in the sequence, the tile to the left the preceding tile, and the tile to the right the next tile.  The tile in the top orange box represents the maximum likelihood prediction of the system.  Each of the three layers consists of 512 LSTM blocks.  All layers are fully connected to the next layer. }
    \label{ANN}
\end{figure*}

These LSTM blocks can be composed in multiple layers with multiple LSTM blocks per layer, and for this work we used 3 internal layers, each consisting of 512 LSTM blocks which can be seen in figure 2.  The input layer to the network consists of a One-Hot encoding where each tile has a unique binary flag which is set to 1 if the tile is selected and all others are 0.  The final LSTM layer goes to a SoftMax layer, which acts as a Categorical probability distribution for the One-Hot encoding.  

Our networks were trained using Torch7 (\cite{TORCH7}) based on code from Andrej Karpathy (\cite{UNREASONABLE_EFFECTIVENESS}).  In addition to choosing size and depth of the network, RNNs (and subsequently LSTMs) come with the additional hyperparameter of how many time steps to consider during the back-propagation through time, which we set to 200 data points.  A common problem in machine learning is that of memorization, where the algorithm exactly memorizes the input data at the expense of being able to generalize to unseen data.  There exist numerous methods to avoid this overfitting, but for this work we used dropout (\cite{DROPOUT}).  Dropout is a technique where during training a pre-specified percentage of LSTM blocks are dropped from the network (along with their corresponding connections) at random for each training instance.  

This has the effect of effectively training an ensemble of networks during the training of one network and reduces the amount of co-adapting that occurs across nodes.  Having decided on the network architecture, we will now get to the core of our data representation and what we mean by a data point.

\subsection{Data Specification}

LSTMs are most commonly used to predict the next item in a sequence. They do this by producing a probability distribution over possible next items and predicting the most likely item.  For this to work, our data must be a sequence.  If we were to consider a format such as the one used by Dahlskog et al. (\cite{NGRAMLEVELS}) we could simply consider a level as a sequence of slices progressing from left-to-right, but this comes with multiple drawbacks:

\begin{itemize}
 \setlength\itemsep{0mm}
\item It can only reproduce vertical slices encountered in the training set - meaning that it could be utterly unable to handle a previously unseen slice
\item This drastically increases the size of the input space as we will see in a second
\end{itemize}
Instead we treat each individual tile of a level from \textit{Super Mario Brothers} as a point in our sequence, that is, like a character in a string.  The tile types we consider are:

\begin{itemize}
 \setlength\itemsep{0mm}
\item \textbf{Solid} - Any tile that is solid and has no other interactions, most commonly ground, giant mushroom, or inert block tiles of which no distinction is made
\item \textbf{Enemy} - Any enemy, again no distinctions are made between enemies
\item \textbf{Destructible Block} - A block that can be destroyed by Super Mario
\item \textbf{Question Mark Block With Coin} - A ?-Block that only contains a coin
\item \textbf{Question Mark Block With Power-up} - A ?-Block that contains a powerup, either a mushroom/flower, a star, or a 1-up
\item \textbf{Coin} - A coin
\item \textbf{Bullet Bill Shooter Top} - The top of a Bullet Bill shooting cannon
\item \textbf{Bullet Bill Shooter Column} - The column that supports the top of the Bullet Bill shooter
\item \textbf{Left Pipe} - The left side of a pipe
\item \textbf{Right Pipe} - The right side of a pipe
\item \textbf{Top Left Pipe} - The top left side of a pipe
\item \textbf{Top Right Pipe} - The top right side of a pipe
\item \textbf{Empty} - The lack of all other tile types
\end{itemize}

Distinctions are made between the variants of ?-Blocks due to the extreme difference in effect that they make in the game.  The vast majority of ?-Blocks contain only coins and if we were to ignore the difference between ?-Blocks we would be losing crucial information about where the most important power-up bearing blocks are located.  Pipes are one of the most important structures in the game. Since previous systems have shown difficulty accurately producing non-gibberish pipes (\cite{SNODGRASS}), we want to consider the different parts of the pipe separately. It is important to know that the left half of a pipe is different from the right half in the sequence, as the latter can never be encountered before the former.  

Each of these tile types can be considered as a character in a string, but a platformer level has 2 dimensions, not just 1, meaning that we have to induce an ordering across the 2 dimensions.  The na{\"i}ve ordering would be left-to-right and most likely bottom-to-top, as the levels progress from left-to-right and there is a higher density of important tiles at the bottom of a level due to the nature of simulated gravity.  However, this sequencing would be highly problematic for most platformers.  Levels in \textit{Super Mario Brothers} are between 200 and 500 tiles in width, which would mean that the LSTM would need to remember for at least that distance what tile it had placed in the previous row.  While LSTMs are good at remembering over longer distances, this would strain the algorithm for no particularly important reason.  

Instead, we consider eight different induced tile orderings determined by answers to the following three questions: 
\begin{itemize}
 \setlength\itemsep{0mm}
\item \textit{Bottom-To-Top} or \textit{Snaking}?
\item \textit{Includes Path Information}?
\item \textit{Includes Column Depth Information}?
\end{itemize}
We now discuss these possible orderings in depth.

\subsubsection{Bottom-To-Top vs. Snaking}

\begin{figure*}[ht]
\centering
    \includegraphics[width=0.8\textwidth]{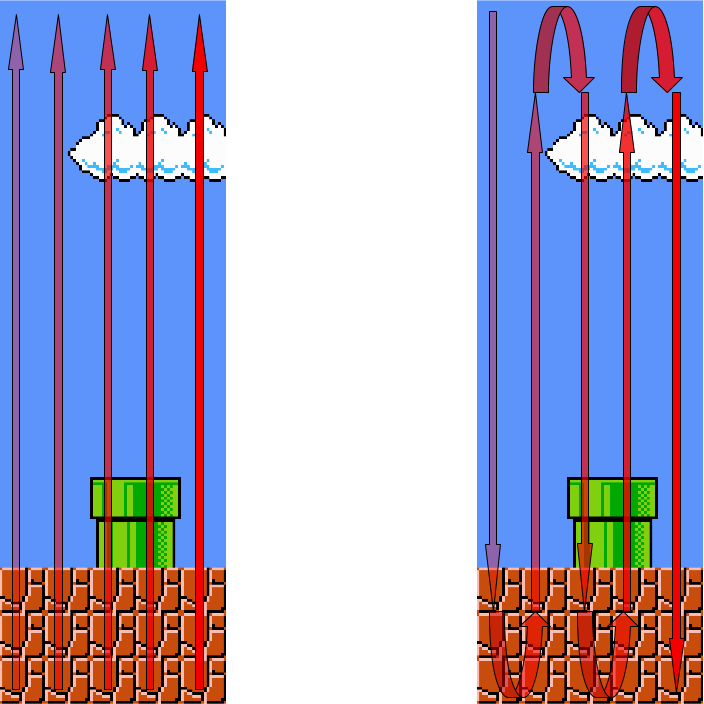}
    \caption{ Illustration of the difference between going from bottom-to-top (LEFT) and \textit{snaking} (RIGHT) for sequence creation}
    \label{fig:snaking}
\end{figure*}
As mentioned, we are considering levels vertically and then horizontally.  This comes first with a decision of whether to go bottom-to-top or top-to-bottom, of which we chose to go from bottom-to-top due to the higher density of important, non-empty tiles at the bottom of a level. This can be seen in the left of figure 3. To do this we include a special character in the sequences to indicate when we have reached the top of the level and are moving to the next column.  

However, there is another, less intuitive manner of progressing through the level: \textit{snaking}.  By snaking we mean alternating bottom-to-top and top-to-bottom orderings as seen in the right of figure 3. 
In this snaking ordering we are following  (\cite{SEQ_TO_SEQ}) which demonstrated that by reversing input streams one can induce better locality in the sequence.  
To see this, consider the pipe in figure 3. In the bottom-to-top ordering there are 17 tiles between the left and right halves of the pipe, while there are only 7 in the snaking ordering.  Snaking also comes with the hidden benefit of doubling the size of our dataset.  When snaking we generate two sequences for a level, one that starts from bottom-to-top as well as one that starts from top-to-bottom. 

\subsubsection{Path Information}

In what we consider to be a novel contribution to the field of machine learned procedural level generation, we also considered the path(s) that a player would/could take through the level.  A common problem with machine learned level generation has been in producing levels that are playable.  Snodgrass and Onta\~{n}\'{o}n found that only roughly half of all levels produced were able to be completed by a simulated agent.  However, if we consider a player's path through the level as a crucial piece of information in the specification of the level and include it in the training information, we are much more likely to generate playable levels since we are not just generating the level geometry but also a player's path through the level.  

We ran a simulated agent through existing levels using a tile-level A$^*$ agent.  Any empty space that the agent passed through was replaced with a special character denoting the player's path.  We considered all paths that were within 10 moves of optimal for this, but this could either be loosened to simulate worse paths or tightened to only consider optimal paths.

\subsubsection{Column Depth}

We also considered how far into the level we are. Obviously, the LSTM has some memory as to where it has been and therefore where it exists in the level, but our history of 200 tiles
means that in essence it only remembers approximately 12 columns in the past.  This captures most patterns that might be encountered, but is unlikely to capture larger considerations such as a ramping in intensity as the player progresses further and further into the level.  To try to get the LSTM to understand level progression we introduced a special character that is incremented every 5 columns.  In other words columns 0-4 have no special character, columns 5-9 have 1 special character, columns 10-14 have 2, etc.

\section{RESULTS}

\begin{figure*}
\centering
    \includegraphics[width=0.5\textwidth]{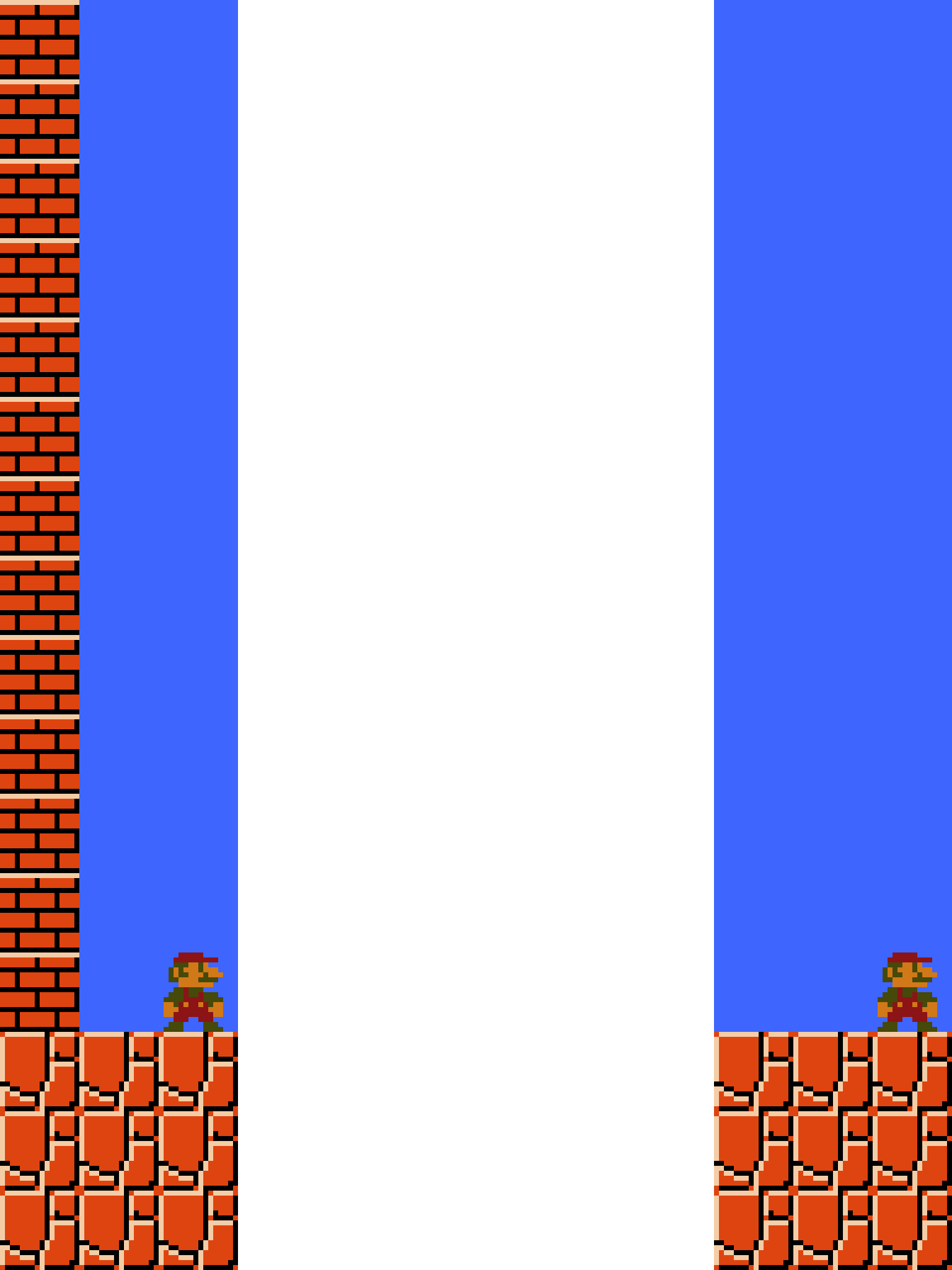}
    \caption{The below ground seed (LEFT) and the above ground seed (RIGHT).}
    \label{fig:seeds}
\end{figure*}

\begin{table}[h]
\begin{center}
\centering
\begin{tabular}{| c | c | c || c |}
\hline
\textbf{Snaking?} & \textbf{Paths?} & \textbf{Line \#?} & \textbf{Negative Log-Likelihood} \\
\hline 
N & N & N & 0.0920 \\
Y & N & N & 0.1540 \\
N & Y & N & 0.0694\\
Y & Y & N & 0.0573 \\
N & N & Y & 0.1376 \\
Y & N & Y & 0.1143 \\
N & Y & Y & 0.0404\\
\textbf{Y} & \textbf{Y} & \textbf{Y} &\textbf{ 0.0177}\\

\hline
\end{tabular}
\end{center}
\caption{The error of the best network for each data format. The \textbf{Snaking-Path-Depth} has the lowest error.}
\label{table:Results}
\end{table} 
We trained eight networks, one for each of the induced sequences. We had a total of 15 levels from \textit{Super Mario Brothers} \nocite{SUPER_MARIO_BROS} and 24 levels from the \textit{Japanese Super Mario Brothers 2} \nocite{SUPER_MARIO_BROS_2} for a total of 39 levels.  We used a 70\%-30\% training-evaluation split in the course of the training.  After every 200 tiles in the training sequence we performed an evaluation on a held out evaluation set, determining how well the LSTM was able to predict unseen sequences. We save this version of the network if it has the best score on the evaluation set, with the belief that this will be the network that has the best ability to generalize from the training set. 
When the training reached a plateau for more than 2 epochs (runs through the training set) we stopped training.  
The error criterion is the negative log-likelihood, which is 0 if the correct value is predicted with 100\% certainty and increases as the tile is less likely to be predicted.

\begin{table*}[h]
\small
\begin{center}
\centering
\hspace{-0.125cm}\begin{tabular}{| c | c | c || c | c | c | c | c | c | c | c | c | }
\hline
\textbf{S?} & \textbf{P?} & \textbf{D?} & $C$ & $e$ & $n$ & $d$ & $p$ & $l$ & $R^2$ & $j$ & $j_i$  \\
\hline  
\multicolumn{3}{| c ||}{Human Authored} & $100\%$ & $ 0.82  $& $ 0.76  $& $ 0.04  $& $ 0.07  $& $ 71.19  $& $ 0.13  $& $ 21.98  $& $ 12.06  $ \\
\hline
N & N & N & $67\% $ & $ \textbf{ 0.81 }$ & $ 0.55$ & $ \textbf{ 0.03 }$ & $ 0.09$ & $ \textbf{ 47.68 }$ & $ \textbf{ 0.15 }$ & $ \textbf{ 20.66 }$ & $ \textbf{ 10.65 }$ \\
Y & N & N & $39\% $ & $ \textbf{ 0.77 }$ & $ 0.52$ & $ \textbf{ 0.04 }$ & $ 0.1$ & $ \textbf{ 33.46 }$ & $ \textbf{ 0.12 }$ & $ \textbf{ 28.61 }$ & $ \textbf{ 6.81 }$ 
 \\
N & Y & N & $96\% $ & $ \textbf{ 0.8 }$ & $ \textbf{ 0.72 }$ & $ \textbf{ 0.04 }$ & $ 0.09$ & $ \textbf{ 59.56 }$ & $ \textbf{ 0.15 }$ & $ \textbf{ 24.17 }$ & $ \textbf{ 12.17 }$  \\ 
Y & Y & N &$93\% $ & $ 0.72$ & $ \textbf{ 0.67 }$ & $ 0.11$ & $ 0.1$ & $ \textbf{ 48.06 }$ & $ \textbf{ 0.19 }$ & $ \textbf{ 19.2 }$ & $ \textbf{ 9.34 }$ \\ 
N & N & Y & $51\% $ & $ \textbf{ 0.82 }$ & $ 0.51$ & $ \textbf{ 0.02 }$ & $ 0.09$ & $ \textbf{ 59.53 }$ & $ \textbf{ 0.17 }$ & $ \textbf{ 19.69 }$ & $ \textbf{ 9.96 }$\\
Y & N & Y& $37\% $ & $ \textbf{ 0.76 }$ & $ 0.44$ & $ 0.09$ & $ 0.1$ & $ \textbf{ 45.99 }$ & $ \textbf{ 0.1 }$ & $ \textbf{ 15.04 }$ & $ \textbf{ 6.68 }$ \\
N & Y & Y& $94\% $ & $ \textbf{ 0.81 }$ & $ 0.64$ & $ \textbf{ 0.03 }$ & $ 0.09$ & $ \textbf{ 46.84 }$ & $ \textbf{ 0.17 }$ & $ \textbf{ 18.31 }$ & $ \textbf{ 8.21 }$ \\
Y & Y & Y & $97\% $ & $ \textbf{ 0.81 }$ & $ \textbf{ 0.68 }$ & $ \textbf{ 0.03 }$ & $ 0.09$ & $ \textbf{ 42.14 }$ & $ \textbf{ 0.19 }$ & $ \textbf{ 19.02 }$ & $ \textbf{ 8.04 }$  \\
\hline
\end{tabular}
\end{center}
\caption{The mean values for each of the considered metrics.  Values in bold are within 1 standard deviation of the original, human-authored levels.}
\label{table:Results2}
\end{table*} 

The results of LSTMs over the evaluation task can be seen in table 1. The change that had the biggest effect on performance was the inclusion of path information, with the networks with paths doing roughly twice as well as the networks without.  Interestingly, the other changes considered without path information worsened network performance, but in tandem with path information improved overall performance. This demonstrates the difficulty of determining a good data specification; each non path change of the data specification away from the na\"{i}ve base case decreased the performance of the LSTM, but when combined with the path information produced performance over twice as good as the next best network and $20 \times$ better than the na\"{i}ve, despite the fact that the most complex specification had a vocabulary that was 36\% larger from the na\"{i}ve specification.  

However, our goal is not to analyze how well a sequence of tiles might have believably come from an existing \textit{Super Mario Brothers} game, but rather to generate new levels with properties hopefully similar to existing \textit{Super Mario Brothers} levels.  To that end we used each of the final trained networks to generate 4000 levels.  Each generator was given a seed sequence and then asked to generate until it reached an end-of-level terminal symbol 2000 times for each of two different seeds, an underground and an  above ground seed as seen in figure 4.  Each seed consists of three columns of tiles, as well as an initial special character denoting the start of the level.   We then analyzed these levels for a number of different properties.  Some, such as linearity and leniency, have been used to evaluate platformer level generation since their inception (\cite{GILLIAN_LINEARITY}).  Others are based off of more recent proposals from Cannosa and Smith (\cite{PCG_EVALUATION}).  In this evaluation we have also included the metrics for the levels from the dataset.  It is hard to generate an intuition for whether a given metric is a good indicator that the generator is producing content that we should be happy with, but since our goal is to produce levels that have properties similar to existing levels we can compare how well the output space of our generator matches that of the existing levels. 
The variants of generators are:
\begin{itemize}
 \setlength\itemsep{0mm}
\item \textbf{S?} - Y if the generator was trained on \textit{snaking} data, N if bottom-to-top
\item \textbf{P?} - Y if the generator had path information, N if not
\item \textbf{D?} - Y if the generator had depth information, N if not
\end{itemize}
The metrics we considered were:
\begin{itemize}
 \setlength\itemsep{0mm}
\item $C$ - The percentage of the levels that are completable by the simulated agent
\item $e$ - The percentage of the level taken up by empty space
\item $n$ - The negative space of the level, i.e. the percentage of empty space that is actually reachable by the player
\item $d$ - The percentage of the level taken up by ``interesting'' tiles, i.e. tiles that are not simply solid or empty
\item $p$ - The percentage of the level taken up by the optimal path through the level
\item $l$ - The leniency of the level, which is defined as the number of enemies plus the number of gaps minus the number of rewards
\item $R^2$ - The linearity of the level, i.e. how close the level can be fit to a line
\item $j$ - The number of jumps in the level, i.e. the number of times the optimal path jumped
\item $j_i$ - The number of meaningful jumps in the level.  A meaningful jump is a jump that was induced either via the presence of an enemy or the presence of a gap.
\end{itemize}
\addvspace{-1\baselineskip}
\begin{figure*}[h]
\centering
\includegraphics[width=1.0\textwidth]{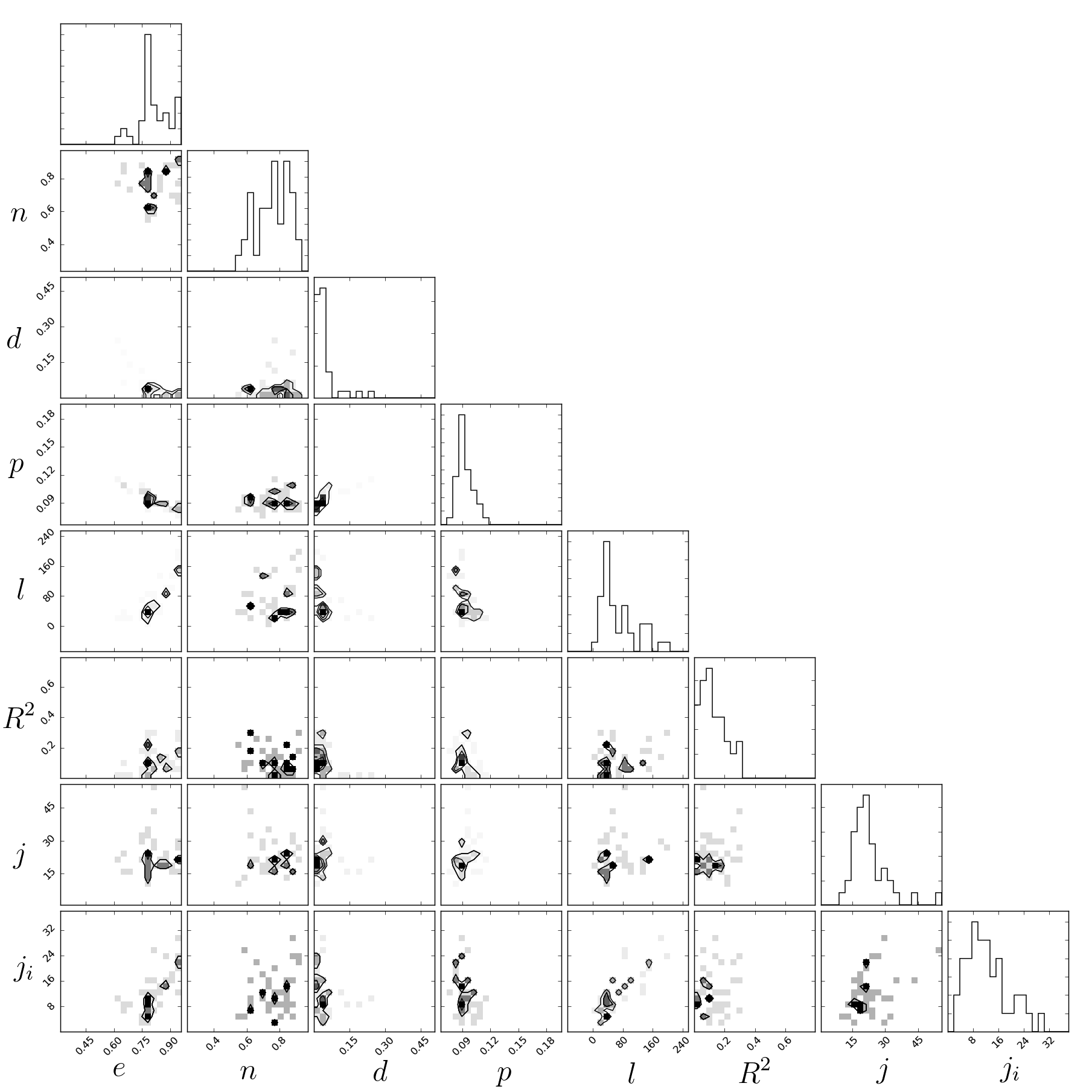}

\caption{ A corner plot of the expressive range for the original levels.  The histograms along the diagonal are for each of the metrics listed along the bottom.  The others are 2D contour plots showing the density of levels with the bottom metric being the X axis and the left metric being the Y axis. }
\label{fig:Originals}
\end{figure*}

\begin{figure*}[h]
\centering
\includegraphics[width=1.0\textwidth]{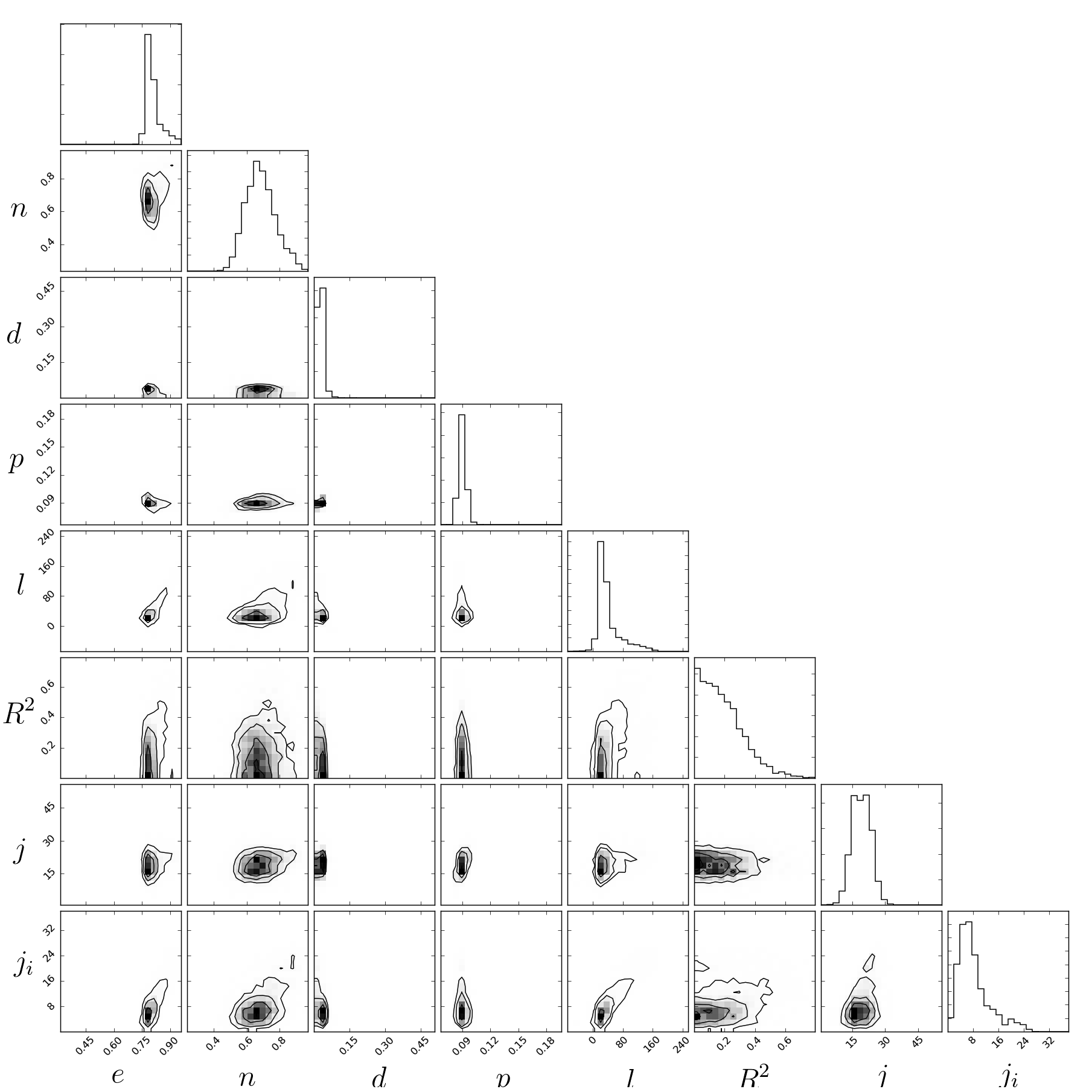}
\caption{ A corner plot of the expressive range for the levels generated by the \textbf{Snaking-Path-Depth} generator. }
\label{fig:SPL}
\end{figure*}

\begin{figure*}[h]
\centering
    \includegraphics[width=1.0\textwidth]{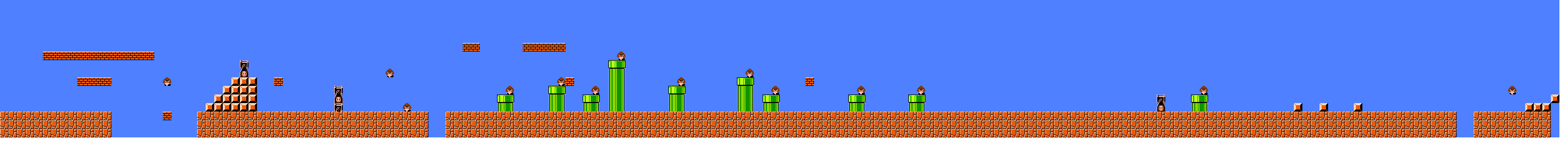}
    
    \includegraphics[width=1.0\textwidth]{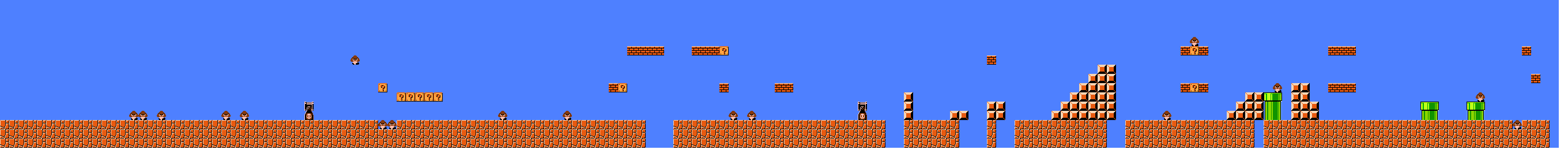}
    
    \includegraphics[width=1.0\textwidth]{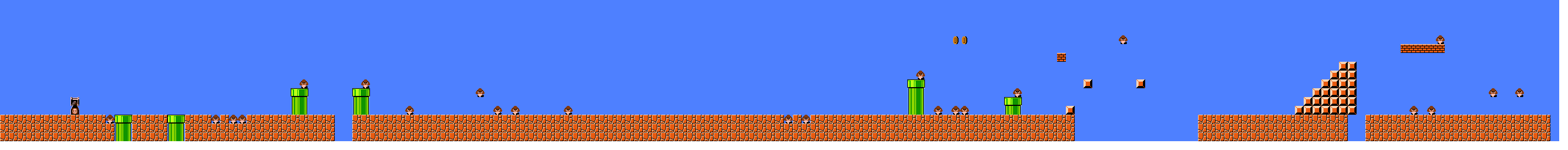}
    
    \includegraphics[width=1.0\textwidth]{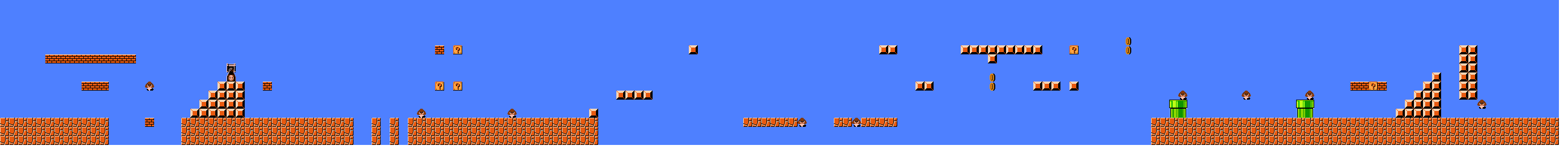}

 \includegraphics[width=1.0\textwidth]{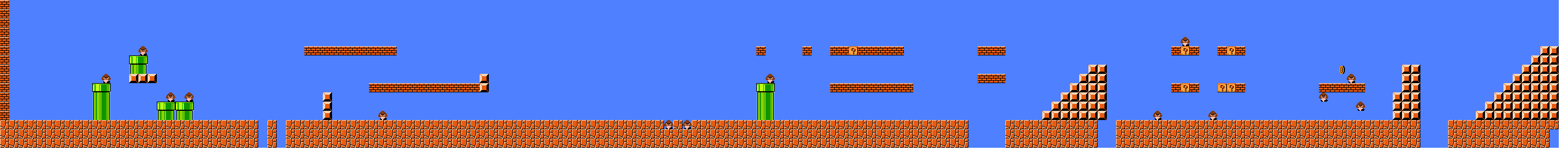}
    \caption{ 5 levels from the \textbf{Snaking-Path-Depth} generator without path information displayed.}
\end{figure*}

\begin{figure*}[h]
\centering
    \includegraphics[width=1.0\textwidth]{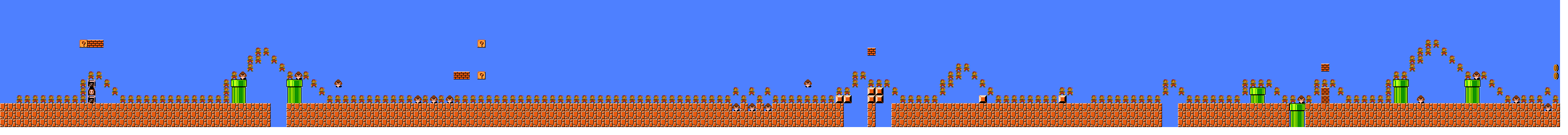}
    
    \includegraphics[width=1.0\textwidth]{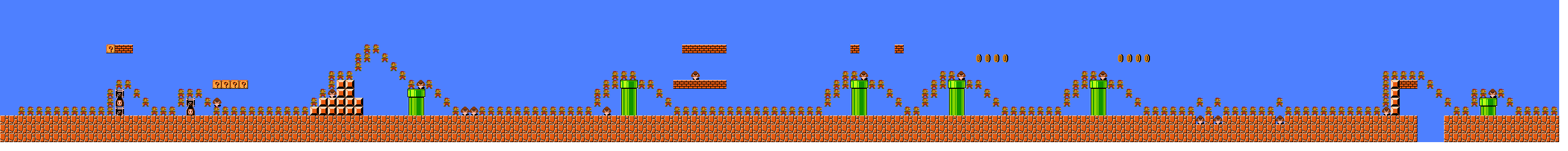}
    
    \includegraphics[width=1.0\textwidth]{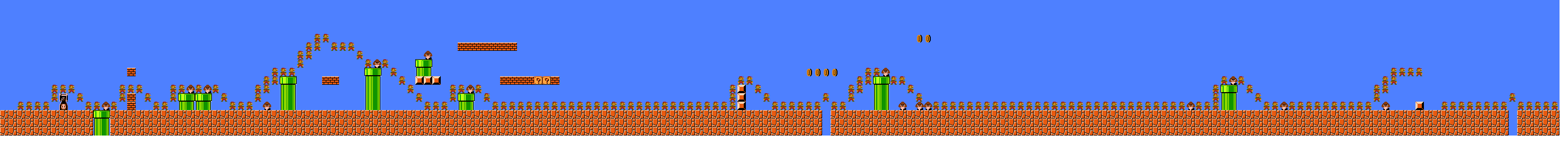}
    
    \includegraphics[width=1.0\textwidth]{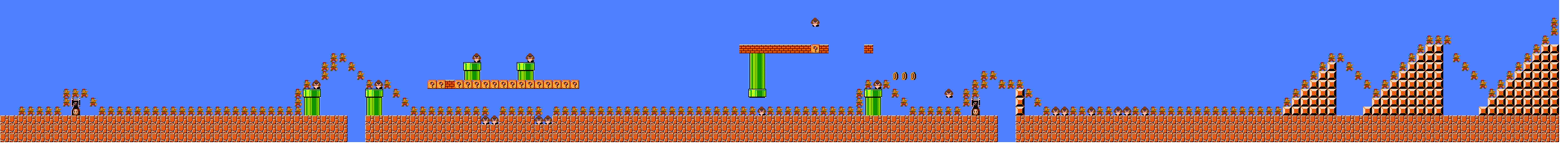}

 \includegraphics[width=1.0\textwidth]{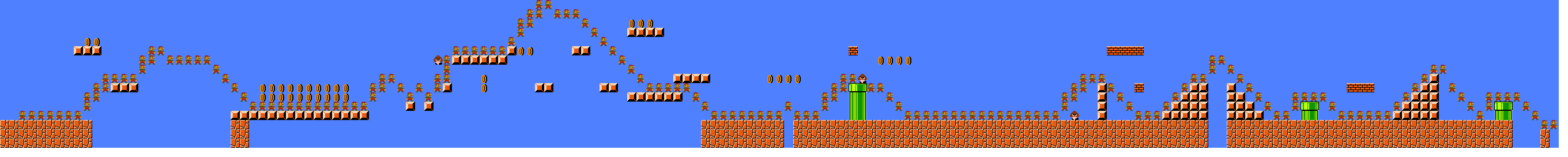}
    \caption{ 5 levels from the \textbf{Snaking-Path-Depth} generator with generated path information displayed.}
\end{figure*}

In table 2 we can see the mean values of the level metrics for the different generators. A metric value is \textbf{bold} if the value was within one standard deviation of the value of the original levels.   All of the generators generally performed  well at matching the existing levels, at least in these metrics.  Certain metrics, like percentage of empty space, leniency, linearity, percentage of decoration, \# of jumps, and \# of induced jumps were 
fairly consistently good across all generators. Interestingly, the lengths of paths through the generated levels were all consistent with each other and all were consistently higher than in the existing levels.

Not surprisingly, the generators with path information did much better at producing completable levels.  To our knowledge, the playability of our generated levels is the best of any generator employing machine learning, beating the best reported of $66\%$ (\cite{HDMC}).  The $97\%$ beats even the best reported human-authored system from the 2011 Mario AI Competition by Mawhorter of $94\%$ as reported by Snodgrass and Onta\~{n}\'{o}n (\cite{HDMC}).

Following are two corner plots which show the density of generated levels across a single dimension (the outer diagonal corresponds to a histogram for the metric along the bottom) or across two dimensions (the left and the bottom corresponding to x and y dimensions).  In figure 4 we can see the original human-authored levels and in figure 5 we can see the results from the \textbf{Snaking-Path-Depth} generator. Generally, the generator does a very good job of matching the expressive range of the original levels..

However, until we are able to train a system to analyze a level for how good it is or how fun it is (which seems a lofty AI-complete goal), such metric-based evaluation must be supplemented with an informal analysis of example generated levels to verify that the metrics actually correspond to human perceptions of high-quality levels. Following are a selection of levels chosen at random from the \textbf{Snaking-Path-Depth} generator.  The first 5 are levels as they would be shown, the latter 5 with the generator's included path information. We believe that these random samples demonstrate the quality of the generated levels and showcase the breadth of patterns it has learned, such as a series of pipes of differing heights, pyramidal structures, and even sequences of unconnected platforms that are still playable. But it is just a random sampling and all of the generated levels can be found online   at http://tinyurl.com/SMBRNN.   

Ignoring the fact that the LSTMs are able to generate playable levels, it is interesting that the LSTMs are even able to produce coherent column-to-column tile placements. Even the fact that the LSTMs never incorrectly produce a column delimiting character is testament to their sequence learning capabilities.  Without fail the LSTMs only produce columns of 16 tiles, exactly the size of the input columns.  This does not come from telling the generator to stop or reset after 16 tiles, but from learned sequences.  

\section{CONCLUSION AND FUTURE WORK}
In this paper we have demonstrated a novel use of LSTM Recurrent Neural Networks for the generation of platformer levels, specifically Super Mario Brothers levels.  Towards this, we have created multiple novel ways of interpreting a Super Mario Brothers level as a linear sequence.  Most importantly, we have demonstrated that the introduction of player path information can dramatically improve the quality of generated levels, most specifically in the playability of generated levels, but also in terms of exhibiting level metrics closer to the human-authored levels. We have also used a wider range of metrics for characterizing our levels than have previously been used, allowing us to perform a more nuanced comparison between human-authored and generated levels.  

We believe that the introduction of path information to the generation process also opens new avenues for automated analysis of levels.  When a designer first designs a level, they often have a hypothesis of what the player will do before any playtesting has occurred. This system works in a similar way by assigning likelihoods to where a player will go based on observation.  We would like to extend this system so that it not only generates levels, but also analyzes presented levels.  While it is possible to send simulated agents through the level to generate analysis, something we ourselves did for this work, future work may enable the system to make suggestions based solely on observations without explicitly modeling or simulating a player. We would like to utilize human playthroughs gathered from gameplay video, to be able to suggest paths that actual human players would take instead of just simple A$^*$ agents.  

Recent work in neural network learning have used attention based systems to learn sequences from data that is not typically thought of as a sequence, such as an image.  While the sequencing in this work is able to produce good results, we feel it could be made more robust by incorporating the sequencing into the learning task.  We also believe that this would enable the system to operate in regimes where a simple left-to-right progression is unsuited, such as games like \textit{The Legend of Zelda} \nocite{THE_LEGEND_OF_ZELDA} or \textit{Metroid} \nocite{METROID}.  These games tend to have highly non-linear progressions with numerous backtrackings and dead-ends required during play.  An attentional system could learn to navigate these non-linear maps.  

A similar concern for future work is deconvolving time and space. In standard platformer levels, the player progresses from left-to-right, which has led to many generators, this one included, to operate as if time has a linear mapping to space, e.g. x-coordinate.  If we were to treat the screen as a tensor with dimensions of width, height, and tile type, we could perform a Tensor auto-regression across time.  For a game like \textit{Super Mario Brothers}, this would work simply as progressing from left-to-right, but for the above mentioned non-linear games, this would work to eliminate the spurious correlation of time and x-coordinate.

\bibliography{bibliography}
\bibliographystyle{aaai}

\end{document}